\documentclass[conference]{IEEEtran}
\IEEEoverridecommandlockouts
\usepackage{multirow,cite}
\usepackage{amsmath,amssymb,amsfonts}
\usepackage{algorithm}
\usepackage{algpseudocode}
\usepackage{graphicx}
\usepackage{textcomp}
\usepackage{xcolor}
\usepackage{tabularx}
\usepackage{booktabs}

\usepackage[square,numbers,sort]{natbib}

\usepackage[normalem]{ulem}
\usepackage{todonotes}

\include{pythonlisting}
\newcommand{\idest}{{\it i.e.}}
\newcommand{\exemp}{{\it e.g.}}
\newcommand{\etc}{{\it etc}}
\newcommand{\etal}{{\it et al.}}

\newcommand{\Tau}{\mathcal{T}}
\newcommand{\bd}[1]{\textbf{#1}}

\def\BibTeX{{\rm B\kern-.05em{\sc i\kern-.025em b}\kern-.08em
    T\kern-.1667em\lower.7ex\hbox{E}\kern-.125emX}}
\begin{document}

\title{Augmented Behavioral Cloning from Observation\\
}

\author{\IEEEauthorblockN{Juarez Monteiro\IEEEauthorrefmark{1}$^1$, Nathan Gavenski\IEEEauthorrefmark{2}$^1$, Roger Granada\IEEEauthorrefmark{1}, Felipe Meneguzzi\IEEEauthorrefmark{3} and Rodrigo Barros\IEEEauthorrefmark{3}}
\IEEEauthorblockA{School of Technology, Pontif{\'i}cia Universidade Cat{\'o}lica do Rio Grande do Sul \\
Av. Ipiranga, 6681, 90619-900, Porto Alegre, RS, Brazil}
\IEEEauthorrefmark{1}\{juarez.santos, roger.granada\}@acad.pucrs.br, \IEEEauthorrefmark{2}nathan.gavenski@edu.pucrs.br \\
\IEEEauthorrefmark{3}\{felipe.meneguzzi, rodrigo.barros\}@pucrs.br
}

\IEEEoverridecommandlockouts
\IEEEpubid{\makebox[\columnwidth]{978-1-4799-7492-4/15/\$31.00~
\copyright2020
IEEE \hfill} \hspace{\columnsep}\makebox[\columnwidth]{ }}

\maketitle
\footnotetext[1]{These authors contributed equally to
the work.}

\begin{abstract}

Imitation from observation is a computational technique that teaches an agent on how to mimic the behavior of an expert by observing only the sequence of states from the expert demonstrations. 
Recent approaches learn the inverse dynamics of the environment and an imitation policy by interleaving epochs of both models while changing the demonstration data.
However, such approaches often get stuck into sub-optimal solutions that are distant from the expert, limiting their imitation effectiveness. 
We address this problem with a novel approach that overcomes the problem of reaching bad local minima by exploring: (i)~a self-attention mechanism that better captures global features of the states; and (ii)~a sampling strategy that regulates the observations that are used for learning. 
We show empirically that our approach outperforms the state-of-the-art approaches in four different environments by a large margin.

\end{abstract}


\section{Introduction}
\label{sec:introduction}

Humans can learn how to perform certain activities by observing other humans. 
This ability of imitating allows humans to transfer the knowledge from demonstrations to the task at hand, despite differences in environment or objects used in the demonstration~\cite{BanduraWalters1977}.
For example, one can learn how to cook by watching videos online, even if the stove and pans are different from the ones in the video. 
The advance in technology and the rising demand for intelligent applications have increased the need for artificial agents that are capable of imitating a human demonstrator. 
Research on imitation learning is motivated by the ease with which humans transfer their knowledge through demonstration rather than articulating it in a way that the interested learner may understand~\cite{RazaEtAl2012}.

Imitation learning, also referred to as \emph{learning from demonstration} (LfD), refers to the task of artificial autonomous agent acquiring skills or behaviors from an expert by learning from its demonstrations~\cite{Schaal1996,ArgallEtAl2009}. 
A natural way of imparting knowledge by an expert is to provide demonstrations for the desired behavior that the learner then emulates~\cite{HusseinEtAl2017}.
Unlike humans that can learn without having direct access to the actions executed in a demonstration~\cite{RizzolattiSinigaglia2010}, classical approaches of LfD use labeled actions in order to imitate the expert behavior.
Such an assumption is restrictive and unrealistic since usually, we do not have direct access to the label of the action that is being performed by the expert. 

Recent approaches perform \emph{imitation from observation} (IfO)~\cite{LiuEtAl2018,TorabiEtAl2018}, which uses only the sequence of state observations from the expert. 
Such approaches learn two models: the inverse dynamics of the environment (Inverse Dynamics Model, IDM) and an imitation policy model (PM). 
Current approaches learn both models iteratively from samples based on each other, \idest, the IDM uses demonstrations generated with a specific policy from PM to update its model, and then the PM is updated using the new outcomes from the updated IDM. 
Using iterations during the learning process allows the policy model to approximate the distribution of actions used by the expert, which improves the imitation process. 
However, performing IfO using this type of iteration has the drawback of overfitting the policy demonstrations, primarily in the first iterations, and sometimes, causing some of the actions to be ignored altogether during learning of the PM due errors in the IDM. 

To deal with this problem in IfO, we design an architecture that uses attention models and a sampling mechanism to regulate the observations that feed the inverse dynamics model, preventing the models from reaching undesirable local minima. 
We name our proposed approach \textit{Augmented Behavior Cloning from Observations} (ABCO). 
It learns a model with the inverse dynamics of the environment in order to infer actions from state changes, and a policy model to mimic the expert via behavior cloning. 
ABCO substantially improves sample efficiency and the quality of the imitation policy model over traditional behavior cloning by exploiting attention mechanisms (Section~\ref{subsec:attention}) within both the inverse dynamic model (Section~\ref{subsec:idm}) and the policy model (Section~\ref{subsec:pm}) and a sampling strategy (Section~\ref{subsec:sampling}) that regulates the observations that will feed the inverse dynamic model. 
Experiments (Section~\ref{sec:experiments}) show that by using either low-dimensional state spaces or raw images as input, ABCO outperforms the main IfO algorithms regarding both \textit{Performance} and \textit{Average Episodic Reward}.

\section{Problem Formulation}
\label{sec:pf}

We formulate the problem of imitation learning within the Markov Decision Process (MDP) framework.
An MDP is a quintuple $M = \{S, A, T, r, \gamma\}$~\cite{SuttonBarto1998}, where $S$ denotes the set of states in the environment, $A$ corresponds to the set of possible actions, $T$ is the transition model $P(s_{t+1} \mid s_{t}, a)$, \idest, a function to determine the probability of the agent transitioning from state $s_t$ to $s_{t+1}$ with $s_{i} \in S$ after taking action $a \in A$ at time $t$; $r$ is a function that determines the immediate reward for taking a specific action in a given state, and $\gamma$ is the discount factor.
The solution for an MDP is a policy $\pi(a \mid s)$ that specifies the probability distribution over actions for an agent taking action $a$ in a state $s_t$ when following policy $\pi$ that imitates the expert behavior.

Since ABCO follows the behavioral cloning from observation (BCO)~\cite{TorabiEtAl2018} framework, we are interested in learning an \emph{inverse dynamics model} $\mathcal{M}_{a}^{s_t,s_{t+1}} = P(a \mid s_t,s_{t+1})$, \idest, the probability distribution of any action $a$ when the agent transitions from state $s_t$ to $s_{t+1}$.
Although we specify the problem as an MDP, the BCO problem is defined without an explicitly-defined reward function, using only \emph{agent-specific states}~\cite{GuptaEtAl2017}, and having no access to the labels of the actions performed by the expert.
Hence, our problem consists in finding an imitation policy $\pi$ from a set of state-only demonstrations of the expert $D = \{\zeta_1, \zeta_2, \ldots, \zeta_N\}$, where $\zeta$ is a state-only trajectory $\{s_0, s_1, \ldots, s_N\}$.

The environment interactions are designed as either pre-demonstrations $\mathcal{I}^{pre}$ or  post-demonstrations $\mathcal{I}^{pos}$, where each demonstration contains a set of \textit{interactions} ($s_t, a_t, s_{t+1}$).
Pre-demonstrations set $\mathcal{I}^{pre}$ contains \textit{golden truth} actions, since the agent takes a random action $a_t$ in state $s_t$ of the environment and generates the new state $s_{t+1}$. 
Conversely, $\mathcal{I}^{pos}$ contains predicted actions from the model since given some state $s_t$ the model predicts action $a_t$ that tries to mimic the expert behavior and generates the new state $s_{t+1}$.


\section{Augmented Behavioral Cloning from Observation}
\label{sec:abco}

Behavioral Cloning from Observation (BCO)~\cite{TorabiEtAl2018} combines both an \emph{inverse dynamics model} to infer actions in a self-supervised fashion, and a \emph{policy model}, which is a function that tells the agent what to do in each possible state of the environment. 
The former considers the problem of learning the agent-specific inverse dynamics, and the latter considers the problem of learning an imitation policy from a set of demonstration trajectories. 
We detail both components, as well as the modifications to this framework we propose in this paper: the \textit{Augmented Behavioral Cloning from Observation} (ABCO) approach.

\subsection{Inverse Dynamics Model}
\label{subsec:idm}

We model the \emph{inverse dynamics model} (IDM) as a neural network responsible for learning the actions that make the agent transition from state $s_t$ to $s_{t+1}$.
In order to learn these actions without supervision, the agent interacts with the environment using a random policy $\pi$, generating pairs of states $\Tau_{\pi_{\phi}}^{ag} = \{(s_{t}^{ag}, s_{t+1}^{ag}), \dots \}$ for agent $ag$ with the corresponding actions $\mathcal{A}_{\pi_\phi} = \{a_{t}, \dots \}$.
We store pairs of states along with their corresponding action ($s_t, a_t, s_{t+1}$) as a pre-demonstration ($\mathcal{I}^{pre}$).
While randomly transitioning from states in $\mathcal{I}^{pre}$, the model learns the inverse dynamics  $\mathcal{M}_\theta$ for the agent by finding parameters $\theta^*$~that best describe the actions that occur for achieving the transitions from $\Tau_{\pi_\phi}^{ag}$.
BCO uses the maximum-likelihood estimation (Eq.~\ref{eq:maxest}) to find the best parameters, where $p_\theta$ is the probability distribution over actions given a pair of states representing a transition. 
At test time, the IDM uses the learned parameters to predict an action $\hat{a}$ given a state transition $(s_{t}^{ag}, s_{t+1}^{ag})$.

\begin{equation}
    \theta^* = \text{arg}\,\max\limits_{\theta}\, \prod_{t=0}^{|\mathcal{I}^{pre}|} p_\theta (a_{t} \mid s_{t}^{\pi_\phi},s_{t+1}^{\pi_\phi})
    \label{eq:maxest}
\end{equation}

We augment the original IDM by adding a \emph{Self-Attention} (SA) module \cite{VaswaniEtAl2017,ZhangEtAl2019} (Section \ref{subsec:attention}), which we use to compensate for the large variation of the samples from $\mathcal{I}^{pre}$ to $\mathcal{I}^{pos}$ in the iterative process.
The self-attention forces the IDM to identify what is essential to learn from each state.
When using SA with images, it can identify which part of the image representation of the state is essential for predicting the correct action.

\subsection{Policy Model}
\label{subsec:pm}

The \emph{Policy Model} (PM) is responsible for cloning the expert's behavior. 
Based on the expert demonstrations $D = \{\zeta_1, \zeta_2, \ldots, \zeta_N\}$, where each demonstration comprises pairs of subsequent states ($s_t^e, s_{t+1}^e$) $\in \Tau^{e}$, (A)BCO uses the IDM to compute the distribution over actions $\mathcal{M}_\theta (s_t^e, s_{t+1}^e)$ and predict action $\hat{a}$ that corresponds to the movement made by the expert to change from state $s_t$ to $s_{t+1}$. 
With the predicted action (self-supervision), the method builds a set of state-action pairs $\{(s_t^e, \hat{a})\}$ corresponding to the action $\hat{a}$ taken in state $s_t$. 
Then this is used to learn the imitation policy $\pi_\phi$ that mimics the expert behavior in a supervised fashion. 

For behavioral cloning, learning an imitation policy $\pi_\phi$ from state-action tuples \{($s_t^e, \hat{a}$)\} consists of finding parameters $\phi^*$ for which $\pi_\phi$ best matches the provided tuples. 
Originally, BCO employs maximum-likelihood estimation following the Eq.~\ref{eq:policy_phi} for finding the best set of parameters $\phi^*$. 
After training the policy network, it performs imitation learning and stores the sequences of states and predicted actions as post-demonstrations ($\mathcal{I}^{pos}$). 

\begin{equation}
    \phi^* = \text{arg}\,\max\limits_{\phi}\, \prod_{t=0}^{N} \pi_\phi (\hat{a_t} \mid s_{t})
    \label{eq:policy_phi}
\end{equation}

Compared to the original BCO, we augment the PM by adding a \emph{self-attention} module~\cite{VaswaniEtAl2017,ZhangEtAl2019}, further detailed in Section~\ref{subsec:attention}.
Unlike in the IDM, we use SA to reduce the state changes during each iteration, since the self-attention module focus on small details and differentiate better all classes given by the IDM. 
The SA also allows the policy to look non-locally at states, helping the model to learn faster for higher dimensional states (\exemp, \emph{Maze} and \emph{Acrobot}) with a more gradual success rate in between iterations.

\subsection{Iterated Behavioral Cloning from Observation}
\label{subsec:ibco}

Torabi \etal~\cite{TorabiEtAl2018} extend the BCO algorithm using the post-demonstration environment interaction to improve both the IDM and the imitation policy.
The improvement, named BCO($\alpha$), where $\alpha$ represents a user specified hyperparameter to control the number of post-demonstration interactions, works as follows.
After learning the imitation policy, the agent executes the environment to acquire new state-action sequences as post-demonstrations ($\mathcal{I}^{pos}$).
These post-demonstrations are then employed to update the IDM, and further on, the imitation policy itself.

The problem of the iterated BCO is that it only uses the set of post-demonstrations to re-train the IDM.
Thus, for those cases in which the policy still does not have good enough predictive performance, the generated set of post-demonstrations will contain misleading actions for specific pairs of states.
Those erroneous actions tend to degrade the predictive performance of the IDM, which leads to degrading the predictions of the policy in a negative feedback loop.
 
In order to deal with these problems, we create ABCO($\alpha$) that iteratively improves over ABCO via a sampling method that weights how much the IDM should learn from pre-demonstrations ($\mathcal{I}^{pre}$) and post-demonstrations ($\mathcal{I}^{pos}$). 
Algorithm \ref{alg:abco_alpha} summarizes the ABCO($\alpha$) training process, where $\Call{trainIDM}{\mathcal{I}^s}$ refers to using $\mathcal{I}^s$ to find a $\theta^{*}$ that best explains the transitions in the demonstration $\mathcal{I}^s$ as in Eq.~\ref{eq:maxest}. 

\begin{algorithm}[t!]
    \caption{ABCO($\alpha$)}
    \begin{algorithmic}[1]
        \State Initialize the model $\mathcal{M}_\theta$ as a random approximator
        \State Initialize the policy $\pi_\phi$ with random weights
        \State Generate $\mathcal{I}^{pre}$ using policy $\pi_\phi$
        \State Generate state transitions $\Tau^{e}$ from demonstrations $D$
        \State Set $\mathcal{I}^s = \mathcal{I}^{pre}$
        \For { $i \gets 0$ to $\alpha$ }
            \State Improve $\mathcal{M}_{\theta}$ by \Call{trainIDM}{$\mathcal{I}^s$}
            \State Use $\mathcal{M}_{\theta}$ with $\Tau^{e}$ to predict actions $\hat{A}$
            \State Improve $\pi_\phi$ by behavioralCloning($\Tau^{e}$, $\hat{A}$)
            
            \For { $e \gets 1$ to $\left|E\right|$ }
                \State Use $\pi_\phi$ to solve environment $e$ 
                \State Append samples $\mathcal{I}^{pos} \gets (s_t, \hat{a}_t, s_{t+1})$
                \If { $\pi_\phi$ at goal $g$}
                    \State Append $v_e \gets 1$
                \Else 
                    \State Append $v_e \gets 0$
                \EndIf
            \EndFor
            \State Set $\mathcal{I}^s =$ \Call{sampling}{$\mathcal{I}^{pre}$, $\mathcal{I}^{pos}$, $P(g \mid E)$, $v_e$}
        \EndFor
    \end{algorithmic}
    \label{alg:abco_alpha}
\end{algorithm}

Unlike BCO(0), which uses only $\mathcal{I}^{pre}$ to train the IDM, ABCO($\alpha$) updates the training data ($\mathcal{I}^{s}$) in every iteration.
As ABCO($\alpha$) does not have any post-demonstration data in the first iteration, $\mathcal{I}^{s}$ receives all data from $\mathcal{I}^{pre}$.
From the second iteration onwards, $\mathcal{I}^{s}$ receives the concatenation of a sample ($\mathcal{I}_{spl}^{pos}$) from $\mathcal{I}^{pos}$ and a sample ($\mathcal{I}_{spl}^{pre}$) from $\mathcal{I}^{pre}$ using a win-loss probability according to the agent capability of achieving the goal for each environment.


\subsection{Sampling}
\label{subsec:sampling}

For every iteration, our sampling strategy creates new training data $\mathcal{I}^{s}$ containing a set of post-demonstrations $\mathcal{I}_{spl}^{pos}$ and a set of pre-demonstrations $\mathcal{I}_{spl}^{pre}$. 
In order to obtain the sample from post-demonstrations ($\mathcal{I}_{spl}^{pos}$), we first select the distribution of actions given a run $E$ in an environment and the current policy $P(A \mid E;\mathcal{I}^{pos})$. 
We consider only successful runs from $\mathcal{I}^{pos}$, \idest, only state-action sequences in which the agent was able to achieve the environmental goal. 
Note that this goal might be a specific state (i.e. such that the last transition in $\mathcal{I}^{pos}$ is in a set of specified states), or avoiding an undesirable state for a fixed number of transitions. 
We infer these goal states from the type of expert demonstration we receive.
We represent it as $v_e$ in Eq.~\ref{eq:expert_distribution}, where $v_e$ is set to 1 if the agent achieves the environmental goal and zero otherwise, and $E$ is the set of runs in an environment. 

\begin{equation}
    \begin{split}
        \label{eq:expert_distribution}
        P(A \mid E;\mathcal{I}^{pos}) = \frac{\displaystyle\sum_{e \in E} v_e \cdot P(A \mid e)}{|E|}
    \end{split}
\end{equation}

The intuition of using the post-demonstration only for successful runs is that if a policy is unable to achieve the environmental goal, then the post-demonstration alone does not close the gap between what the model previously learned with $\mathcal{I}^{pre}$ and what the expert performs in the environment. 
Using only successful runs also gives us a more accurate distribution of the expert since we are only using those distributions that achieved the goal instead of the random distribution that consisted of a balanced dataset. 
By not adding unsuccessful runs to the training dataset, we solve the problem in which BCO($\alpha$) degrades the performance in both models. 
With the distribution of actions from winning executions, we select the sample $\mathcal{I}_{spl}^{pos}$ from those runs, according to the win probability $P(g \mid E)$, \idest, the probability of achieving a goal in an environment, as shown in Eq.~\ref{eq:policy_sampling}. 

\begin{equation}
    \begin{split}
        \label{eq:policy_sampling}
        \mathcal{I}_{spl}^{pos} = (P(g \mid E) \times P(A \mid E,\mathcal{I}^{pos})) \sim \mathcal{I}^{pos}
    \end{split}
\end{equation}

The sample from pre-demonstrations has a complementary size of the post-demonstrations. 
Thus, to create a sample from pre-demonstrations $\mathcal{I}_{spl}^{pre}$, we use the loss probability with a distribution of actions in pre-demonstrations, denoted by $P(A \mid \mathcal{I}^{pre})$, as demonstrated in Eq.~\ref{eq:random_sampling}. 

\begin{equation}
    \begin{split}
        \label{eq:random_sampling}
        \mathcal{I}_{spl}^{pre} = ((1-P(g \mid E)) \times P(A \mid \mathcal{I}^{pre})) \sim \mathcal{I}^{pre}
    \end{split}
\end{equation}

Complementing the training dataset with random demonstrations offers two main advantages. 
First, it helps the model avoid overfitting from the policy demonstrations. 
Second, in the early iterations, when the policy generates only a few successful runs, and the distribution might not be closer to the expert, the training data guarantees exploration by the IDM. 

Using a win-loss probability, we induce the training data to be closer to the expert demonstration than to the random data, which boosts the model capability of imitating the expert. 
In this setting, the more an agent can achieve its goal, the less we want $\mathcal{I}^{s}$ consisting of $\mathcal{I}^{pre}$ and more of $\mathcal{I}^{pos}$. 
It is important to emphasize that our method is only goal-aware since we consider tuples from successful runs in the sample from post-demonstration, and does not use the reward information for learning or optimizing. 
We do not use reward because not all environments have a function for it. 
On the other hand, most agents have a goal that is relatively easy to visually identify by inspecting the last transition, \exemp~the \emph{mountaincar} reaching the flag pole, arriving at the final square at a \emph{maze}, \emph{acrobot} reaching the horizontal line, and the \emph{cartpole} surviving up to 195 steps, as described in Section~\ref{sub:envs}. 

\subsection{Self-attention Module}
\label{subsec:attention}

Self-Attention (SA)~\cite{VaswaniEtAl2017} is a module that learns global dependencies within the internal representation of a neural network by computing non-local responses as a weighted sum of the features at all positions. 
It allows the network to focus on specific features that are relevant to the task at each step and learns to correlate global features~\cite{gatys2016image}. 

ABCO uses the SA module based on the Self-Attention Generative Adversarial Network (SAGAN)~\cite{ZhangEtAl2019}, since it outperforms prior work in image synthesis. 
In SAGAN the self-attention module computes the key $f(x)$, the query $g(x)$ and the value $h(x)$, given a feature map $x$, with convolutional filters with the equations $f(x) = W_fx$, $g(x) = W_gx$ and $h(x) = W_hx$.
We compute the attention map by performing two different steps.
First, we apply Eq.~\ref{eq_sij} with the current key $f$ and query $g$.

\begin{equation}
    \label{eq_sij}
    s_{ij} = f(x_i)^T g(x_j)
\end{equation}

Second, we calculate the softmax function~$\beta_{j, i}$, over the attention module to the $i^{th}$ location when synthesizing the $j^{th}$ region. 
With the attention map $\beta$ and the values $h(x)$ we compute the self-attention feature maps $a = (a_1, a_2, ..., a_N) \in \mathbb{R}^{C \times N}$, where $N$ is the number of feature locations and $C$ is the number of channels, as illustrated in Eq.~\ref{eq_o}.

\begin{equation}
    \label{eq_o}
    a_j = \upsilon \left( \sum_{i=1}^{N} \beta _{j, i} h(x_i) \right), \\ \upsilon(x_i) = W_\upsilon x_i
\end{equation}

In this equation, $W_f$, $W_g$ and $W_h \in \mathbb{R}^{\hat{C} \times C}$ and $W_v \in \mathbb{R}^{C \times \hat{C}}$, where $\hat{C}$ is $C/k$ to reduce the number of features map.
Furthermore, we have the self-attention feature map $a$, which we weigh by $\mu$, a learnable variable initialized as zero.

The SA module in our method minimizes the impact of the constant changes created by the iterations by weighting all features. 
The model is capable of overlooking the potential local noise an agent might create and focus on features that are more relevant for the action prediction.
It also provides smoother weight updates as a consequence of the weighting of all features.
We believe that during early iterations, SA modules will learn with the random policy dataset how to weight each state, and this will later translate in more accurate labeling when $\mathcal{I}^s$ becomes more of $\mathcal{I}^{pos}$ than $\mathcal{I}^{pre}$.

\section{Implementation and Experimental Results}
\label{sec:experiments}

In order to test ABCO, we perform experiments using the environments (Section~\ref{sub:envs}) of the OpenAI Gym~\cite{brockman2016openai} toolkit. 
We developed two networks comprising vector-based environments and image-based environments (Section~\ref{subsec:implementation}) and evaluated the results in terms of \emph{Average  Episodic Reward} (AER) and \emph{Performance} (P) (Section~\ref{subsec:metrics}). 
We describe our findings and a comparison with the state-of-the-art in Section~\ref{subsec:results}. 

\subsection{Environments}
\label{sub:envs}

\begin{figure}[t!]
    \centering
    \includegraphics[width=0.49\textwidth]{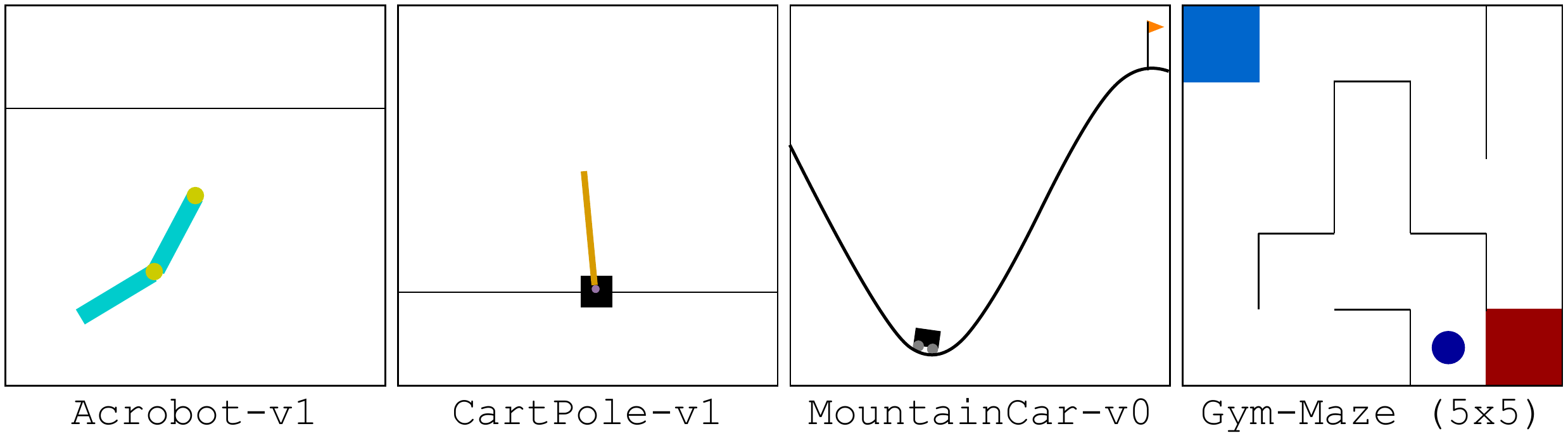}
    \caption{Example frames of Acrobot-v1, CartPole-v1, MountainCar-v0 and Gym-Maze 5x5 environments.}
    \label{fig:envs}
\end{figure}

We perform all experiments using four environments from OpenAI Gym \cite{brockman2016openai}. 
These environments are separated in vector-based environments (\emph{Acrobot-v1}, \emph{Cart-Pole-v1}, \emph{MountainCar-v0}), and image-base environments (\emph{Gym-Maze} $3\times3$, $5\times5$, and $10\times10$). 
Each environment is described below and illustrated in Fig.~\ref{fig:envs}. 

\noindent
\textbf{$\bullet$ Acrobot-v1} is an environment that includes two joints and two links, where the joint between the two links is actuated. 
Initially, the links are hanging downwards, and the goal is to swing the end of the lower link up to a given height. 
The state space contains 6 dimensions: {$\{\cos\theta_1, \sin\theta_1, \cos\theta_2, \sin\theta_2, \theta_1, \theta_2\}$}, and the action space consists of the 3 possible forces. 
Acrobot-v1 is an unsolved environment, \idest, it does not have a specified reward threshold at which it is considered solved. 

\noindent
\textbf{$\bullet$ CartPole-v1} is an environment where an agent pulls a car sideways with the goal of sustaining a pole vertically upward as long as possible. 
The environment has a discrete action space composed of \emph{left} or \emph{right}, while the state space has 4 dimensions: \emph{cart position}, \emph{cart velocity}, \emph{pole angle}, and \emph{pole velocity at tip}. 
CartPole-v1 defines solving as getting average reward of 195 over 100 consecutive trials. 

\noindent
\textbf{$\bullet$ MountainCar-v0} environment consists of a car in an one-dimensional track, positioned between two "mountains". 
The state space has 2 dimensions, the respective car coordinates ($x,y$), and the action space consists of 3 possible strengths to move the car (-1, 0, or 1). 
To achieve the goal in this environment, the car has to acquire the required momentum from the left mountain to drive up to the mountain on the right. 
MountainCar defines solving as getting average reward of -110.0 over 100 consecutive trials. 

\noindent
\textbf{$\bullet$ Gym-Maze} is a 2D maze environment where an agent (the blue dot in Fig.~\ref{fig:envs}, should find the shortest path from the start (the blue square in the top left corner) to the goal (the red square in the bottom right corner). 
Each maze can have a different set of walls configuration, and three different sizes, $3\times3$, $5\times5$, and $10\times10$. 
An agent is allowed to walk towards any wall. 
The agent has a discrete action space composed of \emph{N}, \emph{S}, \emph{W}, and \emph{E}, and the state space consisting of rendered images of the maze.

\begin{table*}[ht!]
  \centering
  \normalsize
  \caption{\emph{Performance} ($P$) and \emph{Average Episode Reward} (AER) for a supervised model (BC), the related work (BCO and ILPO) and our approach (ABCO) using OpenAI Gym environments.}
  \label{tab:results_maze}
  \begin{tabular*}{\textwidth}{c @{\extracolsep{\fill}} c | c c c c c c}
    \toprule
    \multicolumn{1}{c}{Model} & \multicolumn{1}{c |}{Metric} & CartPole & Acrobot & MountainCar & Maze $3\times3$ & Maze $5\times5$ & Maze $10\times10$ \\
    \midrule
    \multirow{2}{*}{BC} 
     & $P$   & \bd{1.000}   & 1.071        & 1.560        & -1.207     & -0.921     & -0.470 \\
     & $AER$ & \bd{500.000} & -83.590      & -117.720     & 0.180      & -0.507     & -1.000 \\
    \midrule\midrule
    \multirow{2}{*}{BCO\cite{TorabiEtAl2018}}
     & $P$   & \bd{1.000}   & 0.980        & 0.948        & 0.883      & -0.112     & -0.416 \\
     & $AER$ & \bd{500.000} & -117.600     & -150.00      & \bd{0.927} & 0.104      & -0.941 \\
    \midrule
    \multirow{2}{*}{ILPO\cite{EdwardsEtAl2019}}  
     & $P$   & \bd{1.000}   & 1.067        & 0.626        & -1.711     & -0.398     & 0.257  \\
     & $AER$ & \bd{500.000} & -85.300      & -167.00      & -0.026     & -0.059     & -0.020 \\
    \midrule 
    \multirow{2}{*}{ABCO($\alpha$)} 
     & $P$   & \bd{1.000}   & \bd{1.086}   & \bd{1.289}   & \bd{1.159} & \bd{0.960} & \bd{0.860} \\
     & $AER$ & \bd{500.000} & \bd{-77.900} & \bd{-132.30} & 0.908      & \bd{0.932} & \bd{0.784} \\
    \bottomrule
\end{tabular*}
\end{table*}

\subsection{Implementation}
\label{subsec:implementation}

We create two different networks to address each type of environment: 
a network for low-dimensional \emph{vector-based environments}, and a network for high-dimensional \emph{image-based environments}. 
We developed all models using the \emph{PyTorch} framework, with the Cross Entropy loss function, and the Adam optimizer~\cite{kingma2014adam}. 
We added self-attention~\cite{VaswaniEtAl2017,ZhangEtAl2019} modules in both IDM and PM. 
Below, we describe the details of each network used in our experiments, where $FC_d$ is a fully connected layer containing $d$ dimensions, $SA_d$ is a self-attention layer, and $\ldots$ indicates the sequence of layers from the original architecture up to the description of the next layer. 

\noindent
\textbf{$\bullet$ Vector-based Environments}: $Input_{dims} \rightarrow FC_{12} \rightarrow SA_{12} \rightarrow FC_{12} \rightarrow SA_{12} \rightarrow FC_{12} \rightarrow FC_{12} \rightarrow Output_{6}$, where \textit{dims} is a vector with twelve and six states for the IDM and PM, respectively. 

\noindent
\textbf{$\bullet$ Image-based Environments}: we modified the ResNet~\cite{HeEtAl2016} architecture by adding two self-attention modules as follows: $Input_{224 \times 224} \rightarrow \ldots \rightarrow ResBlock_{2} \rightarrow SA_{64} \rightarrow \ldots \rightarrow ResBlock_{4} \rightarrow SA_{128} \rightarrow \ldots \rightarrow FC_{dims} \rightarrow LeakyRelu \rightarrow Dropout_{0.5} \rightarrow FC_{512} \rightarrow LeakyRelu \rightarrow Dropout_{0.5} \rightarrow Output_{4}$, where \textit{dims} is a vector with 1024 and 512 features for the IDM and PM, respectively. 

\subsection{Metrics}
\label{subsec:metrics}

We evaluate our policies with two known metrics in the area: the \emph{Average Episodic Reward} ($AER$) and the \emph{Performance} ($P$). 
AER is a standard metric used to measure how well our generated policy performs in a specific environment. 
The metric consists of the average value of one hundred runs for each episode in a given environment (\exemp running one hundred of different mazes in Gym-Maze environment and calculating the average performance, or running one hundred consecutive episodes for the CartPole problem and calculating the average reward). 
AER value is an ideal metric to understand how well the expert did a task and consequently understand how difficult it is to imitate the expert behavior. 

On the other hand, the \emph{Performance} ($P$) metric calculates the average reward for each run scaled from zero to one. 
Zero scores for $P$ represents the reward obtained for a random policy running in a given environment, while the one score represents the reward obtained by the expert policy. 
It is important to note that each environment has its own value for the minimum and maximum rewards, and thus $P$ is not comparable between different environments. 
We do not use accuracy to measure the quality of our generated policies since this metric can not guarantee high-quality results for this problem. 
As mentioned in Section~\ref{sec:abco}, we run our problem in a two-phase approach, where in the first phase, our model can quickly propagate the error to the second. 
Therefore, we cannot use the accuracy as a metric to verify the quality of the generated PM since achieving $100\%$ accuracy with a generated policy using a poor IDM will lead us to lower AER and $P$.

\subsection{Results}
\label{subsec:results}

\begin{figure*}[t]
    \centering
    \includegraphics[width=0.90\textwidth]{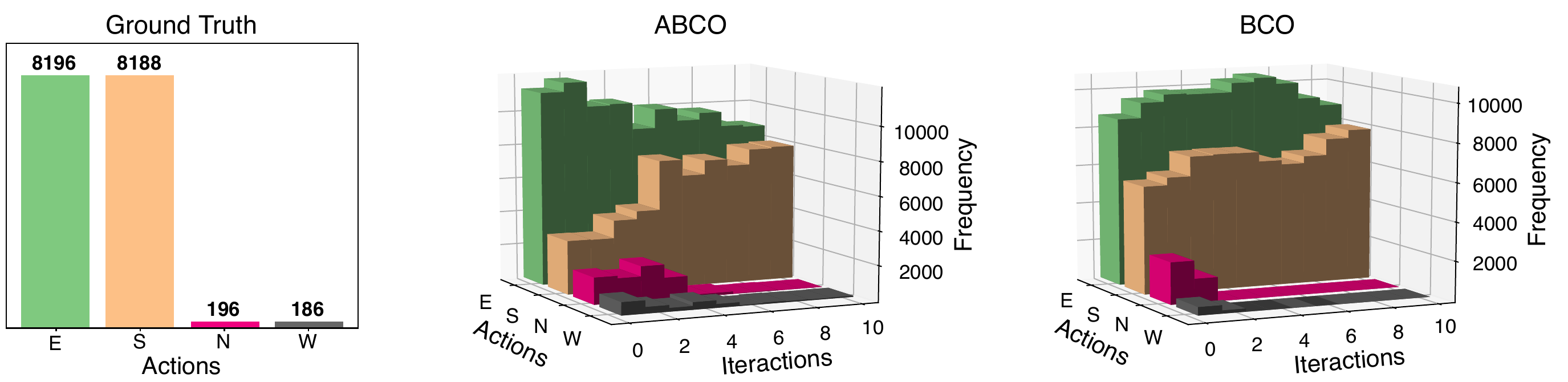}
    \caption{Inverse Dynamic Model predictions of the expert examples through time. It is possible to see that after the first two iterations, due to BCO's Policy poor performance, IDM stopped predicting "North" and "West" classes, while ABCO, although lower than the ground truth, kept predicting sixty and ten after the fifth iteration.
    We believe that the vanishing of the action from BCO is due to all examples from the less present classes in $\mathcal{I}^{pos}$ being worse representations that the random ones, making the Inverse Dynamic model stop predicting those classes, due to expert examples being closer, in the feature space, from other classes than their own.
    }
    \label{fig:distribution}
\end{figure*}


In order to evaluate our approach, we compare our trained models with the state-of-the-art approaches. 
All models are trained using the same initial set of random pre-demonstrations $\mathcal{I}^{pre}$. 
Table~\ref{tab:results_maze} shows the results in terms of \emph{Average Episodic Reward} ($AER$) and \emph{Performance} ($P$) for our models and the related work: BCO~\cite{TorabiEtAl2018} and ILPO~\cite{EdwardsEtAl2019}. 
For comparison purposes, we also show the results for \emph{Behavioral Clone} (BC), which is a supervised approach. 

Table~\ref{tab:results_maze} shows that our method is equal or surpasses the state-of-the-art approaches in all the environments but the Maze $3\times3$ where results are similar to BCO. 
The overall results confirm that the attention module and our sampling strategy can improve the imitation process. 
All approaches achieved the maximum score for CartPole in both $AER$ and \emph{Performance}, showing that this problem is easy to learn. 
Although our model achieved the best $P$ and $AER$ scores in the Acrobot environment, the related work presented similar results with $P\approx1.00$ and $AER=-85.300$. 
Our model achieving better results for $AER$ metric means that ABCO can solve the problem using fewer frames. 
However, both models present similar imitation capabilities since all models achieved $P\approx1.00$. 
It is important to note that even ABCO not using labeled data, it was able to achieve better results than the BC approach that uses labels for actions. 
For MountainCar, we observe a large difference in terms of \emph{Performance}, with our model achieving $P=1.289$ and presenting a difference of $\approx0.34$ to BCO, which is the second-highest result. 

Although in all the Maze environments we achieved the highest scores for \emph{Performance}, we can observe that as the complexity of the environment increases, our performance decreases. 
Nevertheless, we can see in the results that ABCO is less affected than BCO as the complexity increases, since the \emph{Performance} for our results in Mazes $3\times3$, $5\times5$ and $10\times10$ are $1.159$, $0.960$ and $0.860$ respectively, while BCO obtained $0.883$, $-0.112$ and $-0.416$. 
In terms of $AER$, ABCO was only outperformed by BCO in Maze $3\times3$ by $\approx0.02$, where Torabi \etal~\cite{TorabiEtAl2018} achieved $AER=0.927$. 
Comparing the results of ABCO with ILPO, we observe that ILPO increases its \emph{Performance} as the maze increases in size but it is still much lower than ABCO for the $10\times10$ maze.  
We believe that this discrepancy happens for two reasons.
First, our method contains an attention module (\ref{subsec:attention}), which increases the capability of ABCO to focus on essential features through non-visited state spaces. 
Second, ILPO does not consider a full image of the scenario since it uses crop mechanisms and internal manipulations with the state images. 
Using a partial observation from the environment means that the approach cannot receive essential features from the images (\exemp~the initial state, the goal state, the agent, \etc). 
On the other hand, as the maze increases, ILPO receives more local information through the crops, increasing its \emph{Performance}. 

\section{Discussion}
\label{sec:discussion}

We perform an ablation study to observe the impact of each component of our architecture. 
We measure \emph{Performance} and \emph{AER} when using the only the self-attention mechanism without sampling, when using the sampling strategy without self-attention, and when using the combination of attention with different samplings. 
We generate all results using the Maze $5\times5$ environment.  

\begin{table}[!b]
    \centering
    \scriptsize
    \caption{Ablation study considering the 2 main components of ABCO: \emph{attention} and \emph{sampling} in the $5\times 5$ maze environment.}
    \label{tab:ablation}
    \begin{tabular*}{\columnwidth}{@{\extracolsep{\fill}}lrr} 
        \toprule
        Model & $P$ & $AER$ \\ 
        \midrule
        BCO~\cite{TorabiEtAl2018}                 & $-0.112$    &  $-0.941$ \\
        Attention                                 & $-0.415$    &  $-0.940$ \\
        Partial Sampling                          & $0.717$     &  $0.716$ \\ 
        Whole Sampling                            & $0.628$     &  $0.676$ \\
        ABCO (Attention + Partial Sampling)       & $0.960$     &  $0.932$ \\
        ABCO (Attention + Whole Sampling)         & $0.759$     &  $0.755$ \\
        \bottomrule
    \end{tabular*}
\end{table}

\subsection{ABCO and Self-attention}
\label{subsec:abco-self}

To measure the impact on the learning process, we trained ABCO using only the self-attention module. 
We observe that when using only the self-attention, the accuracy of our models was higher than the original method. 
However, high accuracy does not represent an excellent performance since, without the sampling method, some action might not occur in further iterations.
With high accuracy and samples that do not represent the action from the expert accordingly, IDM stops predicting the minority action, creating a sub-set from all possible actions, and the policy learn the new subset of the real actions.
This behavior results in the policy not performing the less frequent actions that are needed to solve different environments during the inference phase (\exemp, North and West, on mazes, or not performing actions during acrobot), as we discuss in Section~\ref{subsec:abco-sampling}.

\begin{figure*}[!ht]
    \centering
    \includegraphics[width=0.9\textwidth]{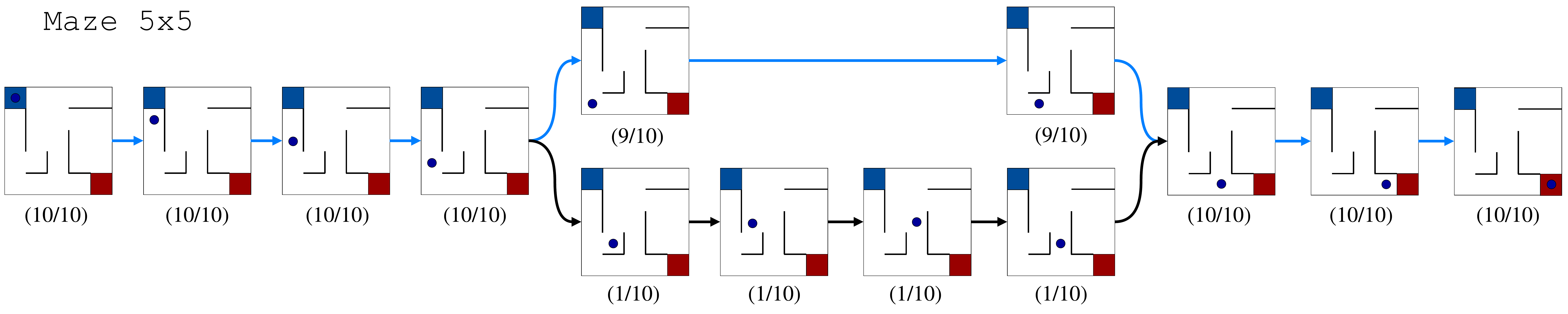}
    \caption{Expert demonstrations executing a 5x5 configuration of Gym-Maze. Bellow the state-image we represent the number of experts that visited that state. The blue line represents the path chosen by our ABCO agent.}
    \label{fig:maze}
\end{figure*}

We observe that even when weighing the features, IDM is still capable of predicting the most common path. 
When feeding the Policy with ten different solutions for each maze, the agent mimics the most common path, as shown in Fig.~\ref{fig:maze}.
Nevertheless, when the first iterations still sample all classes, the model takes more transition samples to reach the results from Table~\ref{tab:ablation}.
Although self-attention alone achieves results similar to BCO($\alpha$), when combined with the sampling method, it has a significant impact on the results.

\subsection{ABCO and Sampling}
\label{subsec:abco-sampling}

In this experiment, we use only the sampling module to train ABCO($\alpha$) by disabling the self-attention module. 
We use the sampling method without the reduction of samples from $\mathcal{I}^{pos}$.
We hypothesize that sampling from the original random policy dataset helps to solve the vanishing actions, as well as close the difference from the first iteration $\mathcal{I}$ and the expert.
The vanishing of actions from the IDM prediction occurs due to the weak policy inference creating a $\mathcal{I}^{pos}$ that does not contain all actions or sparse representations that underfit the inverse dynamic model.
during the early iterations under these conditions, IDM stops predicting the classes that are the minority in the expert dataset. 
This misclassification causes the policy to loop between actions that prevent the model from achieving its goal.
We compare the distribution of all predictions from the IDM from the BCO($\alpha$) and the ABCO in Fig.~\ref{fig:distribution}, where it shows that our sampling method can better predict all classes due to the artificial growth of our dataset caused by sampling from the $\mathcal{I}^{pre}$.

\begin{figure}[b!]
    \centering
    \includegraphics[width=0.48\textwidth]{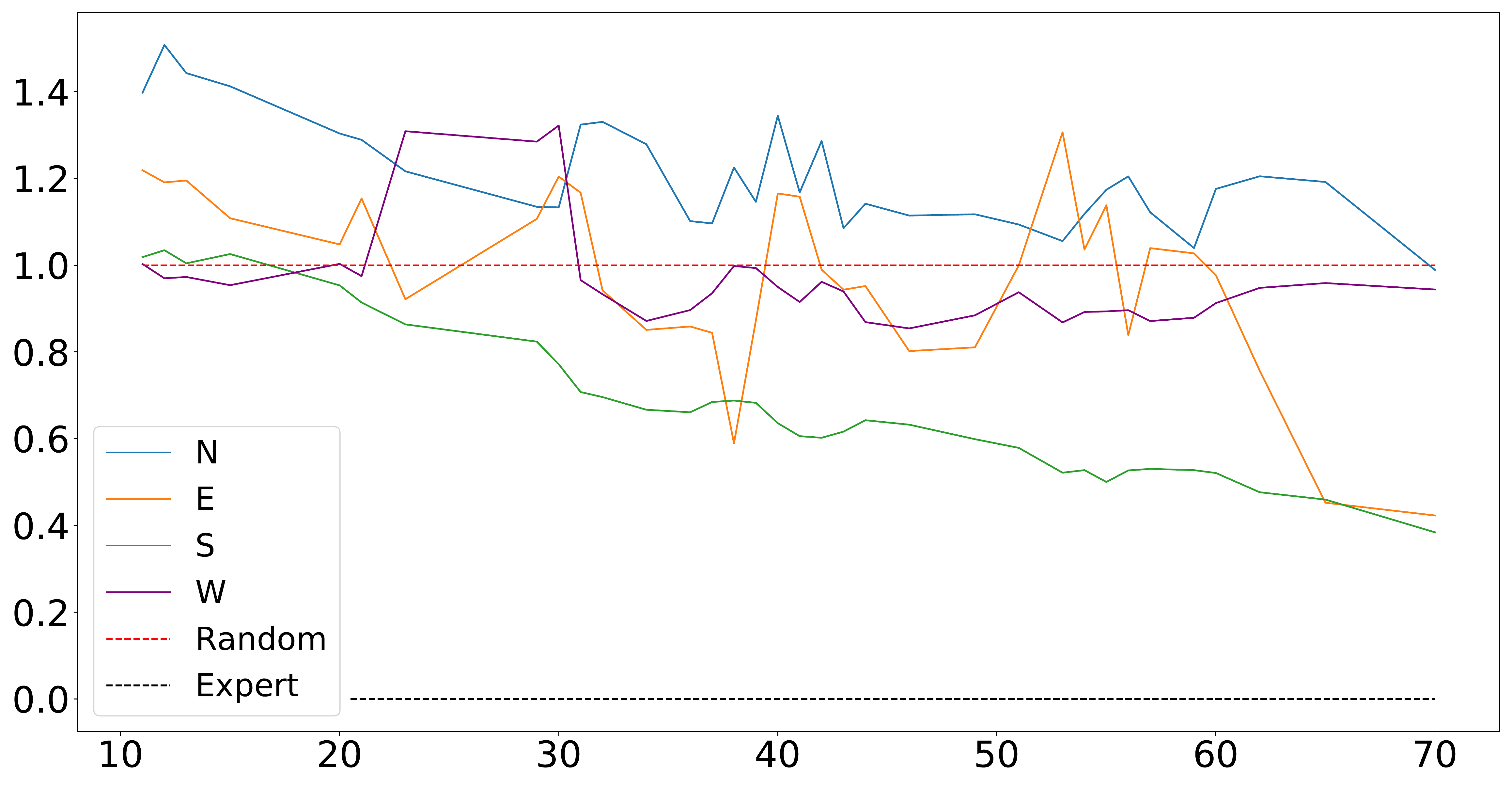}
    \caption{$L2$ distance for the average of each action for each iteration normalized by the expert and random samples in the $5\times5$ mazes.}
    \label{fig:distances}
\end{figure}
Furthermore, to observe if the policy can create samples that are closer from the expert than the random dataset, we calculate the $L2$ distances from the average of all images from each action during each iteration and normalize them between zero, for the expert, and one, for the $\mathcal{I}^{pre}$ samples.
The results in Fig.~\ref{fig:distances} represent how our model learns a policy that creates better $\mathcal{I}$ for the majority classes (\exemp, S and E), and even for the minority classes (\exemp, N and W).
We assume this difference of the approximation of the expert dataset to be due to the minority classes consisting mostly of the $\mathcal{I}^{pre}$ since most mazes do not require those actions.
By sampling from the random dataset, we force our IDM to balance its labeling and create iterations that are further distant. 
Still, as the Policy progresses and solves more runs, it approximates and becomes closer.
By being closer to the expert, the new samples allow the IDM to finetune itself and predict expert labels more precisely.

We also believe that not all interactions following a sub-optimal policy are relevant for IDM's learning.
If our hypothesis is correct that a sub-optimal policy might create samples that harm the IDM ability to label the expert samples correctly, then the values from $AER$ and $P$ would be lower than those from the sampling method from Section~\ref{subsec:sampling}.
In our experiment, we use a Resnet without attention modules and by creating $\mathcal{I}^s$ with all $\mathcal{I}^{pos}$ and the same ratio used in the original sampling method for all $\mathcal{I}^{pre}$.
Using this approach, we observe that when using all interaction from the policy to create the new dataset, the model achieves lower $AER$ and $P$ as expected, as shown in Table~\ref{tab:ablation}.

We conclude that the new sampling method alone can boost the learning experience by allowing the IDM to receive a more balanced dataset. 
Still, when accompanied by the self-attention modules, it improves the generalization from the model by learning to weigh each sample accordingly and further boosting our method performance.

\section{Related Work}
\label{sec:related_work}

Many approaches for imitating from observations have been recently proposed~\cite{HoErmon2016,TorabiEtAl2018,EdwardsEtAl2019,TorabiEtAl2019generative}. 
Ho and Ermon~\cite{HoErmon2016} propose a \emph{generative adversarial imitation learning} (GAIL) approach that learns to imitate policies from state-action demonstrations using adversarial training \cite{GoodfellowEtAl2014}. 

Edwards \etal~\cite{EdwardsEtAl2019} describes a forward dynamics model, \idest, a mapping from state-action pairs \{($s_t$, $a_t$)\} to the next state \{$s_t+1$\}, called \emph{imitating latent policies from observation} (ILPO). 
In their two-step approach, the agent first learns a latent policy offline that estimates the probability of a latent action given the current state. 
Then, in a limited number of steps in the environment, they perform remapping of the actions, associating the latent actions to the corresponding exact actions. 
This approach is very efficient in terms of interactions needed since most of the process occurs offline.

Torabi \etal~\cite{TorabiEtAl2018} develop \emph{behavioral cloning from observation} (BCO) to imitate the behavior of an expert in a self-supervised way by observing its states. 
Their approach contains a model that learns the inverse dynamic of the agent, and a policy model learns which action the agent should use given a state. 
In that work, Torabi \etal~train BCO using only low-dimensional state features.  

Using high-dimensional space, Torabi \etal~\cite{TorabiEtAl2019} explores the fact that agents often have access to their internal states (\idest, \emph{proprioception}). 
In that approach, the architecture learns policies over proprioceptive state representations and compares the resulting trajectories visually to the demonstration data. 

\section{Conclusions and Future Work}
\label{sec:conclusions}

In this paper, we developed a novel approach to learn how to imitate the behavior from experts just by observing their states with no prior information about their actions. 
The pipeline of the architecture includes training two modules iteratively. 
The \emph{Inverse Dynamics Model}, which is responsible for learning actions given that the agent transitioned from two states; and the \emph{Policy Model} that aims to predict which action the agent has to select given a state in order to imitate the expert. 
Using four different environments mentioned in Section~\ref{sub:envs}, we perform experiments showing that our approach can use low-dimensional or raw image data to learn how to imitate an expert, and achieve better results than the best current methods. 

As future work, we aim to evaluate our technique in more challenging domains, such as Continuous Control Tasks, Atari Games, Robotics Goal-based Tasks and others.
Such domains, have larger space states when compared with the ones in our evaluation, which makes them harder to imitate.
We plan to work with temporal approaches and possibly adversarial techniques in order to search for a model that could be able to imitate and generalize having comparable or better results than the state-of-the-art.

\section*{Acknowledgment}
This study was financed in part by the Coordenação de Aperfeiçoamento de Pessoal de Nível Superior - Brasil (CAPES) - Finance Code 001, and CAPES/FAPERGS agreement (DOCFIX 04/2018) process number 18/2551-0000500-2.  
We gratefully acknowledge the support of NVIDIA Corporation with the donation of the graphics cards used for this research.

\bibliographystyle{IEEEtran}
\bibliography{ijcnn}

\end{document}